\documentclass[journal]{IEEEtran}

\usepackage{fancyhdr}            % Permits header customization. See header section below.
\fancypagestyle{plain}{
    \lhead{}
    \fancyhead[R]{\thepage}
    \fancyhead[L]{}
    
    \fancyfoot{}
}
\ifCLASSINFOpdf

\else

\fi

\usepackage{algorithm}
\usepackage{algorithmic}
\usepackage{amsmath}
\usepackage{tikz}
\usepackage{multirow}

\usepackage{booktabs}
\usetikzlibrary{chains,positioning}
\usepackage{float}
\usepackage{zref-xr,zref-user}
\zexternaldocument*{xr02}
\usepackage[caption = false]{subfig}
\usepackage{multicol}
\usepackage{graphicx} % more modern
\usepackage{placeins}
\usepackage{csquotes}
\usepackage{mathrsfs}
\usepackage[font=small,labelfont=bf]{caption}
\usepackage{url}
\usepackage{blindtext}

\usepackage{times}
\usepackage{epsfig}
\usepackage{graphicx}
\usepackage{amsmath}
\usepackage{amssymb}

\usepackage[utf8]{inputenc} % allow utf-8 input
\usepackage[T1]{fontenc}    % use 8-bit T1 fonts
\usepackage{hyperref}       % hyperlinks
\usepackage{url}            % simple URL typesetting
\usepackage{booktabs}       % professional-quality tables
\usepackage{amsfonts}       % blackboard math symbols
\usepackage{nicefrac}       % compact symbols for 1/2, etc.
\usepackage{microtype}      % microtypography
\usepackage{graphicx}
\usepackage{booktabs} % for professional tables

\usepackage{amsmath}
\usepackage{amssymb}
\usepackage{epstopdf}
\usepackage{subfig}

\DeclareMathOperator*{\argmin}{arg\,min}

\begin{document}
\thispagestyle{empty}

\title{DL-Reg: A Deep Learning Regularization Technique using Linear Regression}

\author{Maryam~Dialameh, %~\IEEEmembership{Member,~IEEE,}
        Ali~Hamzeh, %~\IEEEmembership{Fellow,~OSA,}
        and~Hossein~Rahmani%~\IEEEmembership{Life~Fellow,~IEEE}% <-this % stops a space
\thanks{M. Dialameh is with the Department
of Computer Science, Shiraz University, Shiraz,
IRAN, e-mail: \href{mailto:4tiamo4@gmail.com}{4tiamo4@gmail.com}.}% <-this % stops a space
\thanks{A. Hamzeh is with the Department
of Computer Science, Shiraz University, Shiraz,
IRAN, e-mail: \href{mailto:ali@cse.shiraz.ac.ir}{ali@cse.shiraz.ac.ir}.}% <-this % stops a space
\thanks{H. Rahmani is with the School of Computing and Communications, Lancaster University, UK, e-mail: \href{mailto:h.rahmani@lancaster.ac.uk}{h.rahmani@lancaster.ac.uk}.}}

\markboth{IEEE TRANSACTIONS ON NEURAL NETWORKS AND LEARNING SYSTEMS}%
{Shell \MakeLowercase{\textit{et al.}}: Bare Demo of IEEEtran.cls for IEEE Journals}

\maketitle
\thispagestyle{empty}
\begin{abstract}
Regularization plays a vital role in the context of deep learning by preventing deep neural networks from the danger of overfitting. This paper proposes a novel deep learning regularization method named as DL-Reg, which carefully reduces the nonlinearity of deep networks to a certain extent by explicitly enforcing the network to behave as much linear as possible. The key idea is to add a linear constraint to the objective function of the deep neural networks, which is simply the error of a linear mapping from the inputs to the outputs of the model. More precisely, the proposed DL-Reg carefully forces the network to behave in a linear manner. This linear constraint, which is further adjusted by a regularization factor, prevents the network from the risk of overfitting.
The performance of DL-Reg is evaluated by training state-of-the-art deep network models on several benchmark datasets. The experimental results show that the proposed regularization method: 1) gives major improvements over the existing regularization techniques, and 2) significantly improves the performance of deep neural networks, especially in the case of small-sized training datasets. 
\end{abstract}


\IEEEpeerreviewmaketitle

\section{Introduction}
\label{intro}
Regularization is a popular and essential technique in the field of machine learning, reducing the complexity of learned models while allowing them to predict accurately over a set of unseen data during the test phase \cite{R1}. This technique even becomes more vital in the concept of deep learning because of their highly nonlinear behaviors, which adding even one more layer increases the nonlinearity of models to a considerable extent \cite{R2}, resulting in a poor generalization performance on unseen samples.

According to a taxonomy proposed in \cite{R3}, deep learning regularization techniques could be divided into five main categories. Data-based regularization is the first category, which aims to either simplify the representation of input data by applying certain transformations or creating a large number of data points using data augmentation techniques. The second category is based on modifying the network structure such as imposing restrictions on the number of nodes/layers and choosing a proper activation function. Regularization via the error function is the third type of this taxonomy trying to add certain features to the error function such as robustness to imbalanced data. The fourth category is based on modifying the optimization algorithm used to learn network. Termination methods, DropOut, momentum, and weight initialization are some examples of this category. The last category is based on adding a regularization term into the network loss-function. It is assumed that there is no dependency between the regularization term and targets in this category, and they are independent of each other. Weight decay and $\ell_2-norm$ are two typical examples of this category. 

Because of the highly nonlinear behavior of deep learning models, especially when they become deeper by adding more layers, they naturally tend to do more memorization than generalization. This problem even becomes more serious when the size of train-data is small. Moreover, although many researchers have tried to address this problem, it undoubtedly needs more work, as the danger of overfitting has remained not completely solved in the context of deep learning models. Moreover, none of the current existing regularization methods explicitly try to enforce deep networks to behave less nonlinear. That is, there is no method yet that explicitly penalizes deep networks from learning a highly nonlinear model, and this area is still open to research. Additionally, it is desirable for a regularizer to be efficient, simple, computationally inexpensive, and result in discriminative features maps. Satisfying all these characteristics together, however, is subject to more research.

Considering the aforementioned points, this paper proposes a simple but efficient regularization method, named as DL-Reg (an abbreviation for Deep Learning Regularization), by adding a regularization term into the network\textquoteright \thinspace s loss function, enforcing the learned model to explicitly behave as much linear as possible. In other words, the proposed DL-Reg, which could be categorized into the last category outlined above, not only explicitly penalizes the network from learning a purely highly nonlinear model, but also gives enough motivation for learning as linear as possible while preserving the discrimination ability of the model. To accomplish this, we take the advantage of linear regression and propose a least-squares error-term, representing the squared error of a linear mapping from the inputs of the network to its outputs. This term simply motivates the network to behave as linearly as possible so that minimizes the least-squares errors. Additionally, the main contributions of this work can be summarized as follows:
\begin{description}
	\item[$\bullet$ ] Regularizing deep networks using the sum of squared errors of a linear regression model which directly maps the inputs to the output of the network
	\item [$\bullet$ ] Allowing supervised deep networks to be trained using the semi-supervised learning
	\item[$\bullet$ ] Increasing the performance of deep networks on the small-sized dataset
	\item[$\bullet$ ] Conducting extensive experiments to certify the significance of the proposed method in enhancing the performance of deep networks
\end{description}
The rest of this paper is structured as follows. Section \ref{sec2} provides a brief survey of the related regularization approaches. Section \ref{sec3} presents the proposed regularization method in depth. Section \ref{sec4} reports the experimental results, and section \ref{sec5} discusses the findings and provides several possible future trends. Finally, Section \ref{sec6} concludes the paper.

\section{Related Work}
\label{sec2}
This section covers a brief background of several regularization methods used in the context of deep learning. $\ell_2-norm$, which works similar to Weight Decay in the case of SGD-optimizer \cite{R34}, is perhaps one of the well-known traditional regularizing methods, which is simply $\frac{\gamma}{2}||w||^2_2$, where $\gamma$ is a regularization factor, and $w$ is the weights of the network \cite{R2,R21}. In a recent work \cite{R19}, however, it was shown that separating the weight decay from the gradient-based updating rule can substantially improve the generalization ability of the learning, particularly in the case of Adam optimizer. 

Smoothness~\cite{R4} is another regularization method, which penalizes large derivatives in the model and is defined by $||Jac_{f_w}(x)||_F$, where $Jac(.)$ and  $||.||_F$ denote the Jacobian of the network $f$ parametrized by $w$ and the Frobenius norm, respectively. In another work, $\ell_2-norm$ of the gradient of loss function was applied to obtain a loss-invariant backpropagation, which makes the loss invariant to the input changes \cite{R5}. Hessian Penalty \cite{R6} has been proposed as a fast approximation of $\ell_2-norm$ of the Hessian of the network by penalizing Jacoobian with noisy inputs. This idea was further exploited in \cite{R25} to build a robust network against adversarial examples.

To improve the performance of recurrent neural networks (RNN), it is shown that imposing unitary or orthogonal constraints on the weight matrices prevents the network from the problem of vanishing/exploding gradients \cite{R7,R8}. In another research, matrix spectral norm~\cite{R9} has been used to regularize the network by making it indifferent to the perturbations and variations of the training samples. More precisely, the parameters of the model are trained so that the spectral norm of weights is small, allowing the network to not be sensitive to the changes in the order of training data at each epoch. In the same direction, SHADE~\cite{R20} has been proposed whose loss function is based on the conditional entropy trying to minimize the variation in the input representations. Inspired by \cite{R23}, Louizos \textit{et al.} \cite{R22} leverage the notion of weight sparsity, trying to set a certain number of the weights of the network as zeros by applying $\ell_0-norm$; however, it is applicable for a certain condition, as $\ell_0-norm$ is not generally differentiable. In the same way, group sparse regularization method~\cite{R24} applies the notion of  $\ell_{2,1}-norm$ sparsity on the sets of outgoing weights from neurons.

Shake-Shake regularization~\cite{R10} was proposed for only a specific type of residual network (ResNet). It follows the idea of adding gradient noise to the learning procedure where gradient noise is replaced by gradient augmentation, allowing the network to escape from local optima. More precisely, this approach multiplies the output of residual branches by a random scaler and adds the results to both forward and backward passes. ShakeDrop regularization~\cite{R11} is an extension of Shake-Shake that can be applied to other ResNet models. 

The family of drop methods, initially introduced by DropOut~\cite{JMLR:v15:srivastava14a}, is another type of regularization method, which prevents deep learning techniques from overfitting by randomly dropping a certain number of neurons during different epochs of training. DropBlock \cite{R29}, DropBand \cite{R30}, DropFilter \cite{R31} are several recently proposed methods of this family. DropBlock randomly drops a certain number of continuous regions in feature maps. DropBand drops one channel of input data each time, and DropFilter randomly drops some elements of convolutional layers. Additionally, Spectral DropOut~\cite{KHAN201982} is another member of this family, which prevents overfitting by firstly calculating Fourier coefficients of the network and then eliminating noisy and weak coefficients; it, however, needs additional calculations to find Fourier coefficients. Cutout \cite{R26} is a restricted type of DropOut-based regularizers  that randomly masks squared regions of inputs. This masking forces the network to learn complementary features, which is helpful in case of occlusion. Overall, having a closer look at this family, one can see that such methods are different in terms of dropping layers and/or dropping nodes/weights; however, there is not much difference between them in terms of performance in practice, and they behave almost like DropOut. 

Moosavi-Dezfooli \textit{et al.}~\cite{R27} proposed a new regularization technique by minimizing the curvature of the loss surface, which is helpful in the case of adversarial robustness. However, their optimization procedure needs to calculate the eigenvalues of a Hessian loss, which is computationally complex. Stankovic \textit{et al.}~\cite{R33}, took the advantage of graph Laplacian regularizer to address the problem of limited training data in deep neural networks. The proposed method is based on iteratively solving a quadratic programming problem, which adds more computation to the training phase. Apart from that, this method works only for binary classification problems.

A modification of the softmax loss function, which is called Angular softmax~\cite{R28}, was recently proposed as an explicit regularization technique, trying to increase the inter-class separability by distancing between class centers. Although this method, theoretically, leads to more discrimination, the empirical results over different types of datasets are far from expectations. Moreover, Angular softmax is a new/modified loss function, not a regularization method in general. Another recent work is style transfer regularization~\cite{R32}, which tries to regularize the network by generating new data, mostly textured image data, through combining the content of an image with the appearance of another one.

Based on the above summary, we can conclude that none of the current regularization methods have all the properties of simplicity/generality, efficiency, and dealing with small-sized training datasets at the same time. Accordingly, this paper aims to propose an efficient, but simple, method for regularizing deep neural networks, allowing the networks to extract highly discriminative features, as the experimental results certify this assertion. Moreover, the proposed method is suitable for the case of small-sized training datasets. 

\section{Proposed Method}
\label{sec3}
This section describes the proposed regularization method (DL-Reg) in details. A deep neural network can be considered as a function parameterized by a set of weights that maps an n-dimensional input $x \in \mathfrak{R}^n$ to a c-dimensional output $f_\mathcal{W}(x)\in \mathfrak{R}^c$, i.e., $f_\mathcal{W}  \colon x \rightarrow f_\mathcal{W}(x)$. The goal of training is to find an optimal set of weights $\mathcal{\mathcal{W}}^*$ minimizing a certain empirical risk function $J(\mathcal{W})$:
\begin{equation}
\label{e1}
\begin{array}{l}
\mathcal{W}^* =  \underset{\mathcal{W}} \argmin  J(\mathcal{W}; X,Y),
\end{array}
\end{equation}
where $X=[x_1, x_2, \cdots, x_m]^T \in \mathfrak{R}^{m \times n}$ and $Y=[y_1, y_2, \cdots, y_m]^T \in \mathfrak{R}^{m \times c}$  is the corresponding binary label matrix. $Y$ is defined as follows: for each training sample $x_i (i=1,\cdots,m)$, $y_i \in \mathfrak{R}^c$ is its label vector. If $x_i$ is from the $k$th class $(k=1,\cdots,c)$, then only the $k$th entry of $y_i$ is one and all the other entries are zero. The risk function $J$ then takes the following form:
\begin{equation}
\label{e2}
\begin{array}{l}
J(\mathcal{W}; X,Y) =  \sum_{i=1}^m L(f_\mathcal{W}(x_i), y_i) + \gamma \Omega (...),
\end{array}
\end{equation}
where $L$ is a loss function, which calculates the error between the output of the network $f_\mathcal{W}(x_i)$ and the target $y_i$, $\Omega$ is a regularization function that may consume a certain number of inputs except the targets, and $\gamma$ is a regularization factor, which determines the importance of regularization in the risk function. Considering the aforementioned definitions, this work aims to propose a regularization function $\Omega$, which improves the generalization ability of the network, particularly in case of small sample size problems. Accordingly, we propose the following regularization function, which is simply the squared norm of error between a linear mapping of the inputs and the outputs of the network: 
\begin{equation}
\label{e3}
\begin{array}{l}
\Omega_Z(X,f_\mathcal{W}(X)) = ||\dot{X}Z - f_\mathcal{W}(X)||_2^2,
\end{array}
\end{equation}
where $Z \in \mathfrak{R}^{(n+1)\times c}$ is a linear transformation operator (the last row of $Z$ represents bias parameters), which maps $\dot{X} \in R^{m \times (n+1)}$ (i.e. $X$ concatenated by a column of all ones) to the c-dimensional output $\dot{X}Z$, and $||.||_2$ denotes $\ell_2-norm$. In other words, $\Omega_Z$ calculates the error of a linear regression between the inputs and the outputs of the network. The parameters of $Z$ are initialized randomly and then updated during the training process of the deep neural network. More precisely, whenever the parameters $\mathcal{W}$ of the network get updated, $Z$ is updated as well. 

\begin{figure}[H]
	\vskip 0.1in
	\includegraphics[width=1\linewidth]{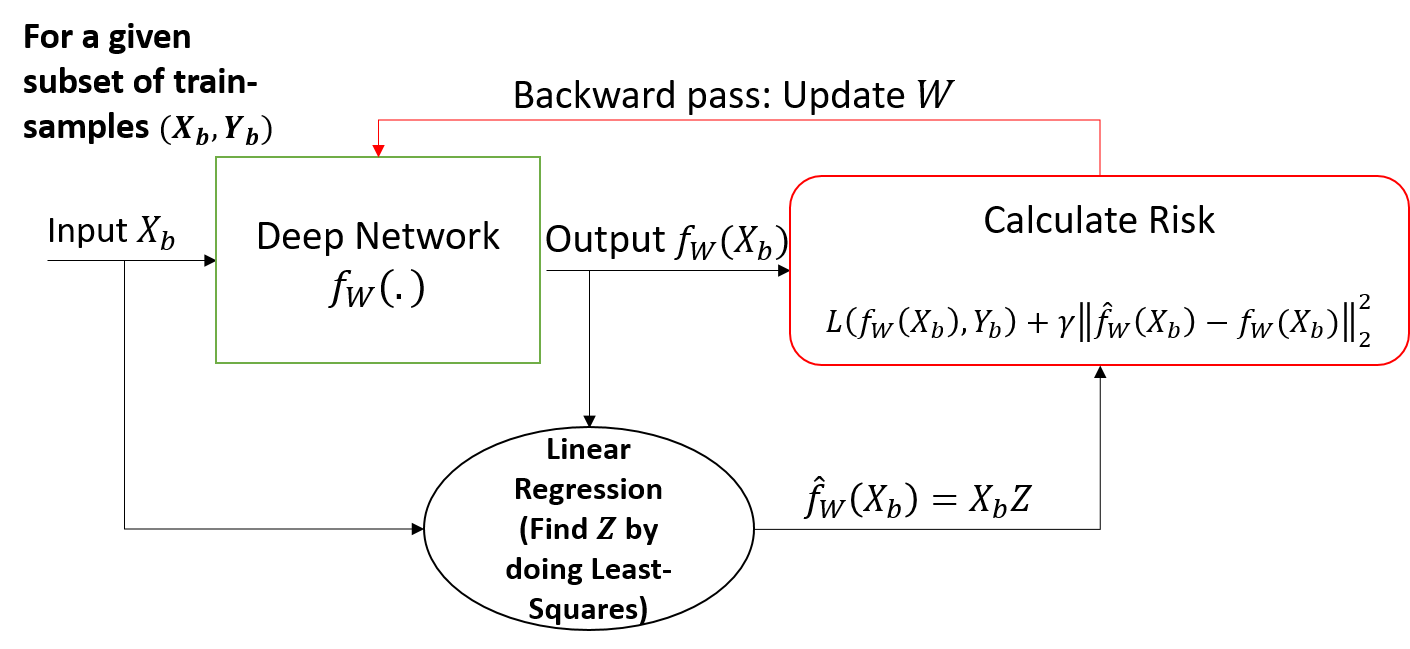}
	\caption{A schematic overview of the proposed regularization method.}
	\label{mf1}
	\vskip 0.0in
\end{figure}

Practically speaking, Eq. (\ref{e3}) can be applied for the case of mini-batch optimization. Therefore, it can be rewritten as follows:
\begin{equation}
\label{e4}
\begin{array}{l}
\Omega_Z(X_b,f_\mathcal{W}(X_b)) = ||X_bZ - f_\mathcal{W}(X_b)||_2^2
\end{array}
\end{equation}
where $X_b\in \mathfrak{R}^{s\times (n+1)}$ represents a mini-batch subsamples of $X$ concatenated by a column of all ones as the biases multipliers, $s$ is the size of a batch, and $f_\mathcal{W}(X_b)\in \mathfrak{R}^{s\times c}$. Figure \ref{mf1} shows a graphical view of DL-Reg.\\

Minimizing Eq. (\ref{e4}) w.r.t. $Z$ is a typical least-squares problem and could be solved by a closed-form solution as follows:
\begin{equation}
\label{e5}
\begin{array}{l}
Z^* = X_b^T(X_bX_b^T)^{-1}f_\mathcal{W}({X_b})
\end{array}
\end{equation}

Because the number of samples in a mini-batch, $s$, is often  smaller than the size of the input, $n$, \textit{i.e.}, $s \ll n$, $X_b$ is a fat matrix, and can be accounted as a full-row rank matrix. Hence, the inverse of $X_bX_b^T$ exists as it forms a full rank matrix.

It is also worthwhile noting that in a case of $s\gg n$, which is almost quite rare in deep learning problems, Eq. (\ref{e4}) could then be solved as follows:
\begin{equation}
\label{e6}
\begin{array}{l}
Z^* = (X_b^TX_b)^{-1}X_b^Tf_\mathcal{W}({X_b}),
\end{array}
\end{equation}
in which $X_b$ becomes a full-column rank, \textit{i.e.}, tall matrix, and consequently $(X_b^T X_b )^{-1}$ exists \cite{R12}.\\

Eq. (\ref{e3}) aims to keep the network to behave as a linear mapping function and penalize the network when it behaves highly nonlinear. Hence, one concern about the proposed $\Omega$-function might be its negative impact on the nonlinearity power of the deep networks, as the major power of deep learning methods is rooted in their abilities to produce nonlinear feature maps. This concern, however, can be rejected because the regularization factor $\gamma$ adjusts the impact of linearization enforced by $\Omega$-function, and choosing a proper value of $\gamma$ can easily resolve this concern. It is worthwhile noting that the parameter $\gamma$ could be selected through a cross-validation procedure over a validation set. Additionally, the independence of $\Omega$-function from the targets allows the network to take the advantage of unlabeled training data and makes it also suitable for the case of semi-supervised learning \cite{R13}.

\section{Experiments}
\label{sec4}
This section evaluates the effectiveness of the proposed regularization method on several state-of-the-art deep network architectures, such as ResNet-152 \cite{R14}, DenseNet \cite{R16}, and VGG \cite{R17}. The evaluation is performed on the task of image classification using several benchmark datasets including MNIST, CIFAR-10, CIFAR-100, and ImageNet. Needless to say that the parameters used in the training phase of each architecture, such as learning rate, epochs, and batch-size are the same for both cases of using and not using our proposed regularization method.

\subsection{CIFAR DATASETS}
\begin{figure*}[t]
	
	\subfloat[\scriptsize DensNet-121-Train]{\includegraphics[width = \dimexpr\columnwidth/2\relax]{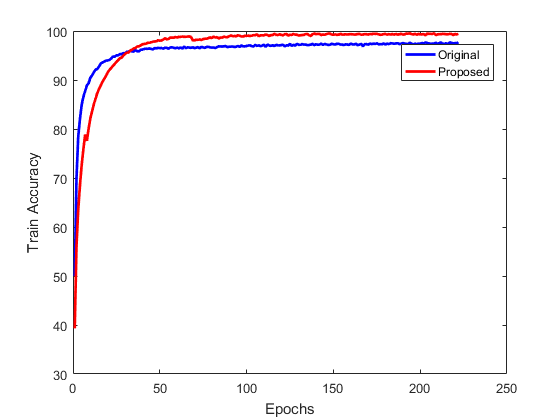}}
	\subfloat[\scriptsize VGG13-Train]{\includegraphics[width = \dimexpr\columnwidth/2\relax]{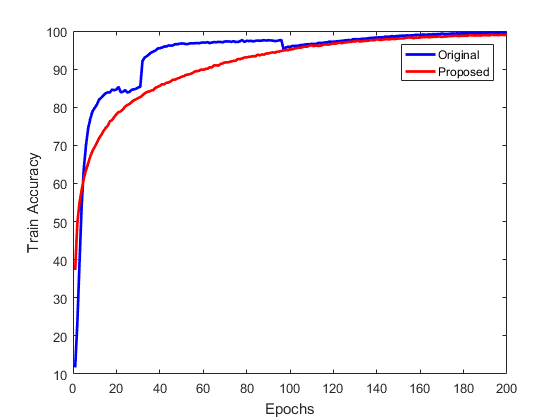}}
	\subfloat[\scriptsize ResNet-Train]{\includegraphics[width =  \dimexpr\columnwidth/2\relax]{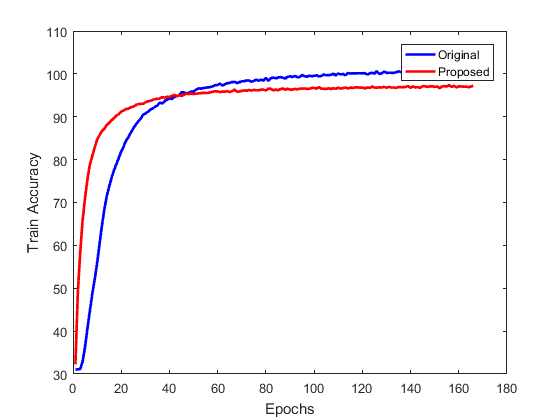}}
	\subfloat[\scriptsize EfficientNetB0-Train]{\includegraphics[width =  \dimexpr\columnwidth/2\relax]{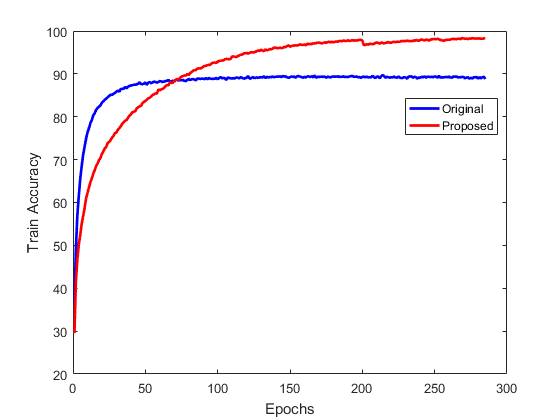}}\\
	
	\subfloat[\scriptsize DensNet-121-Test]{\includegraphics[width =  \dimexpr\columnwidth/2\relax]{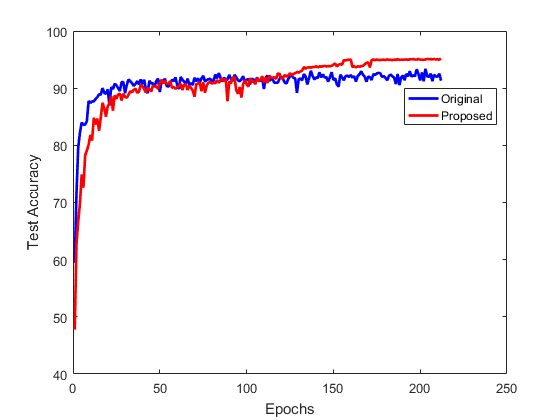}}
	\subfloat[\scriptsize VGG13-Test]{\includegraphics[width =  \dimexpr\columnwidth/2\relax]{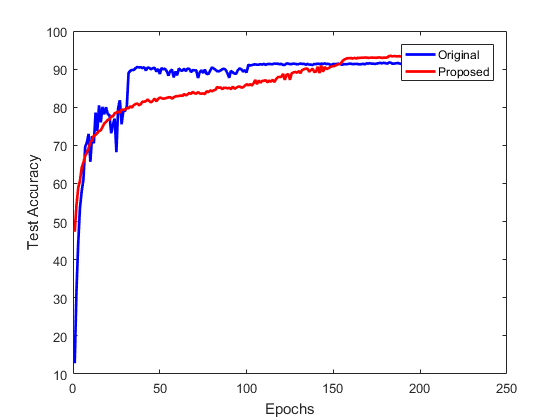}}
	\subfloat[\scriptsize ResNet-Test]{\includegraphics[width =  \dimexpr\columnwidth/2\relax]{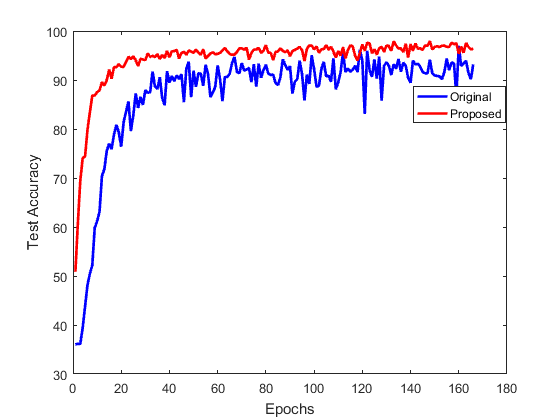}}
	\subfloat[\scriptsize EfficientNetB0-Test]{\includegraphics[width =  \dimexpr\columnwidth/2\relax]{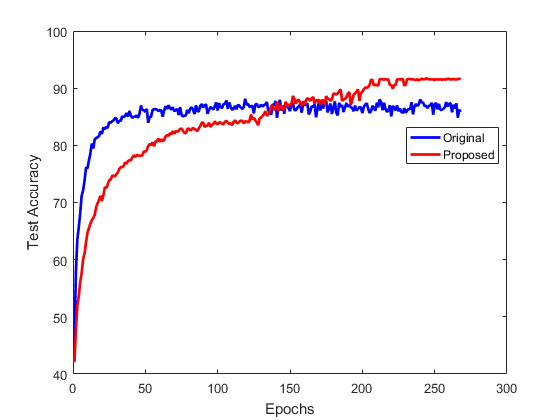}}
	\caption{CIFAR10- The classification accuracies are obtained by applying our proposed regularization method on four different networks. Rows show training/testing results obtained from each network.}
	\label{f1}
\end{figure*}

\begin{figure*}[t]
	\subfloat[\scriptsize DensNet-121-Train]{\includegraphics[width = \dimexpr\columnwidth/2\relax]{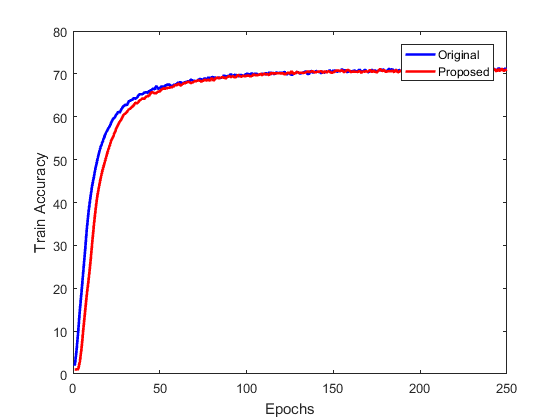}} 
	\subfloat[\scriptsize VGG13-Train]{\includegraphics[width =  \dimexpr\columnwidth/2\relax]{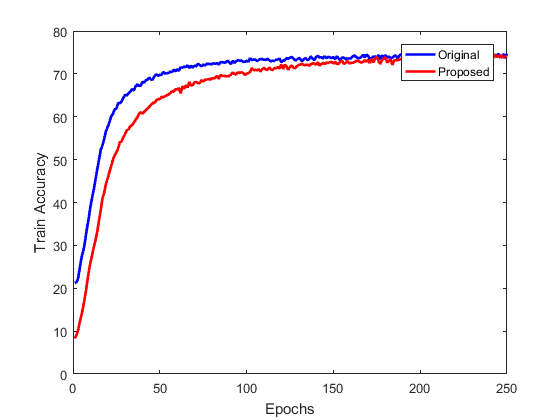}}
	\subfloat[\scriptsize ResNet-152-Train]{\includegraphics[width = \dimexpr\columnwidth/2\relax]{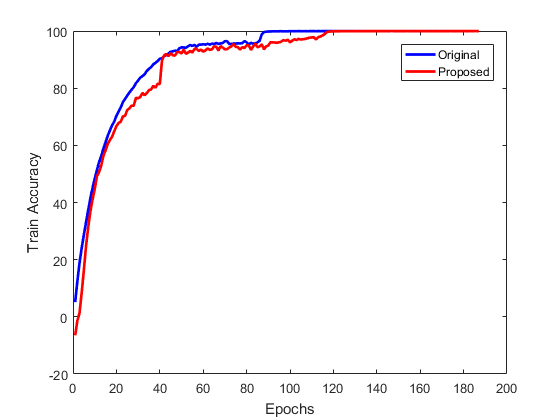}}
	\subfloat[\scriptsize EfficientNetB0-Train]{\includegraphics[width =  \dimexpr\columnwidth/2\relax]{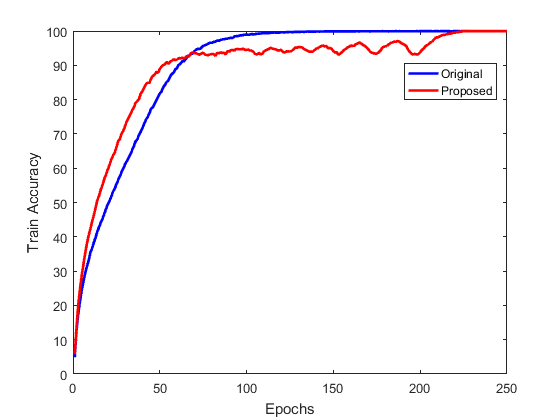}}\\
	
	\subfloat[\scriptsize DensNet-121-Test]{\includegraphics[width =  \dimexpr\columnwidth/2\relax]{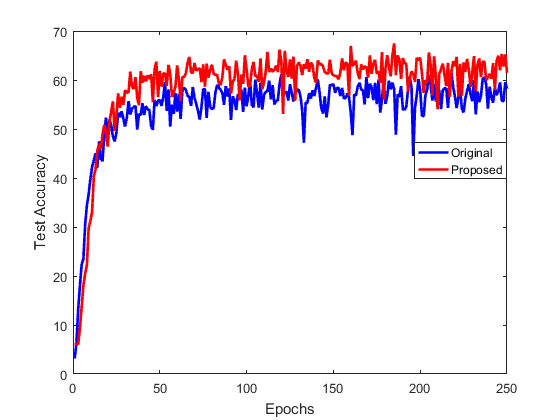}}
	\subfloat[\scriptsize VGG13-Test]{\includegraphics[width =  \dimexpr\columnwidth/2\relax]{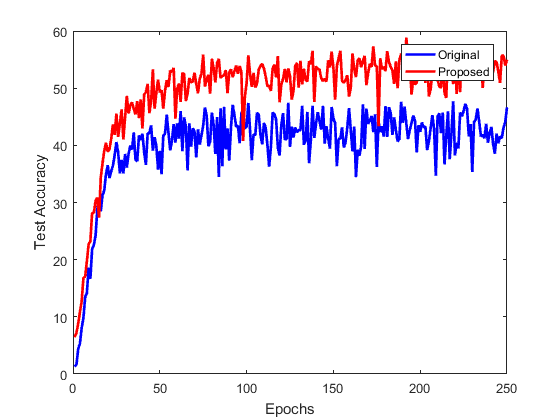}}
	\subfloat[\scriptsize ResNet-152-Test]{\includegraphics[width =  \dimexpr\columnwidth/2\relax]{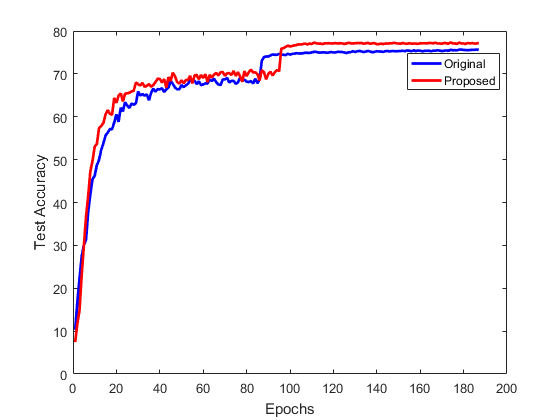}}
	\subfloat[\scriptsize EfficientNetB0-Test]{\includegraphics[width = \dimexpr\columnwidth/2\relax]{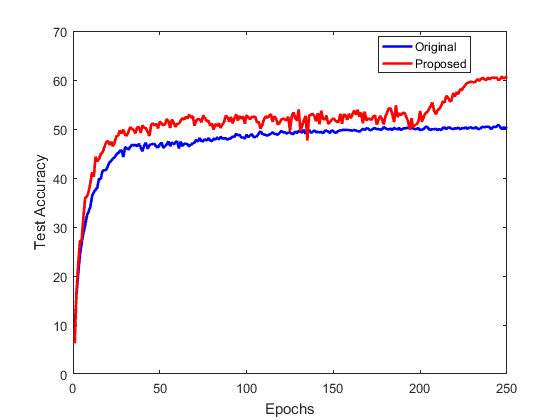}}
	\caption{CIFAR100- The classification accuracies are obtained by applying our proposed regularization method on four different networks. Rows show training/testing results obtained from each network.}
	\label{f2}
\end{figure*}

\begin{figure*}[t]
	
	\subfloat[\scriptsize DensNet-121-Train]{\includegraphics[width = \dimexpr\columnwidth/2\relax]{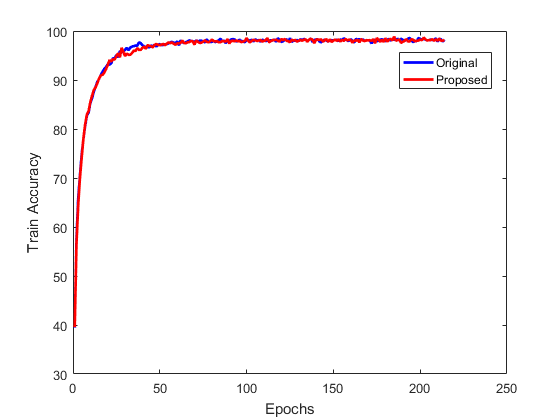}}
	\subfloat[\scriptsize VGG13-Train]{\includegraphics[width =  \dimexpr\columnwidth/2\relax]{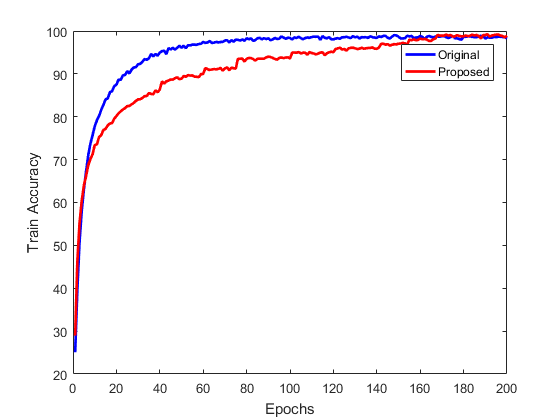}}
	\subfloat[\scriptsize ResNet152-Train]{\includegraphics[width = \dimexpr\columnwidth/2\relax]{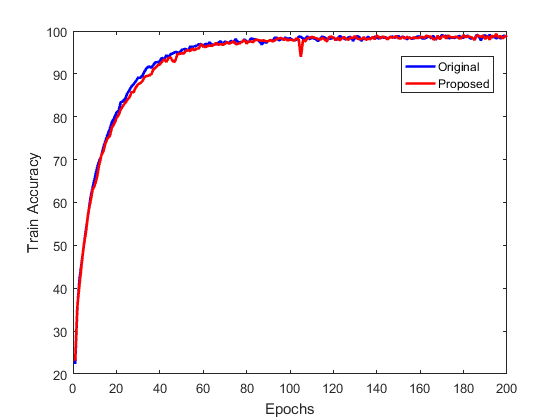}}
	\subfloat[\scriptsize EfficientNetB0-Train]{\includegraphics[width =  \dimexpr\columnwidth/2\relax]{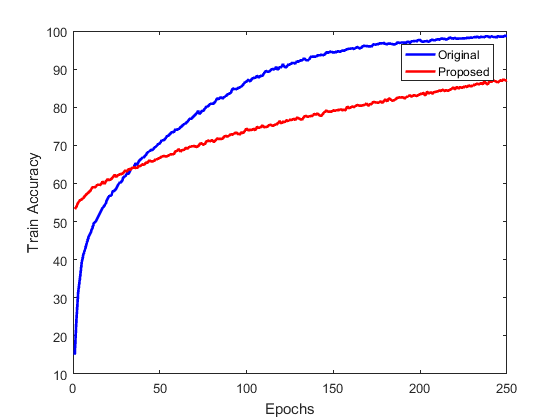}}\\
	
	\subfloat[\scriptsize DensNet-121-Test]{\includegraphics[width = \dimexpr\columnwidth/2\relax]{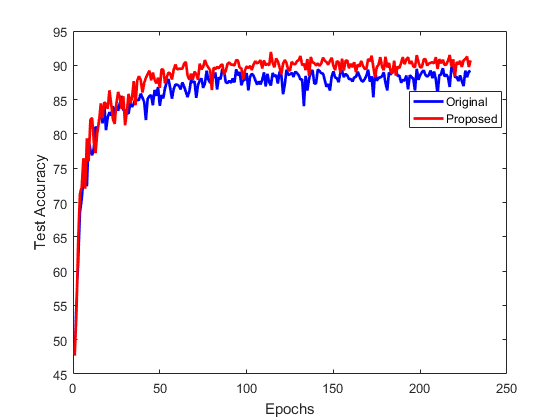}}
	\subfloat[\scriptsize VGG13-Test]{\includegraphics[width =  \dimexpr\columnwidth/2\relax]{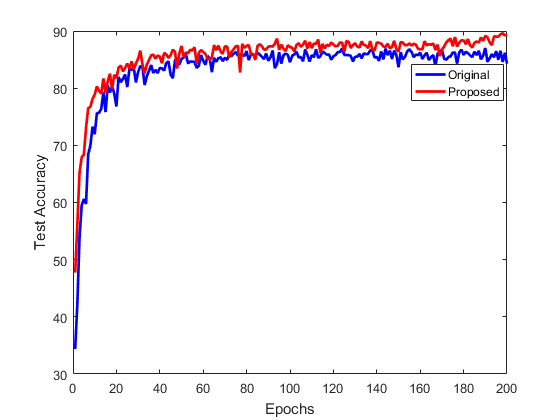}}
	\subfloat[\scriptsize ResNet152-Test]{\includegraphics[width =  \dimexpr\columnwidth/2\relax]{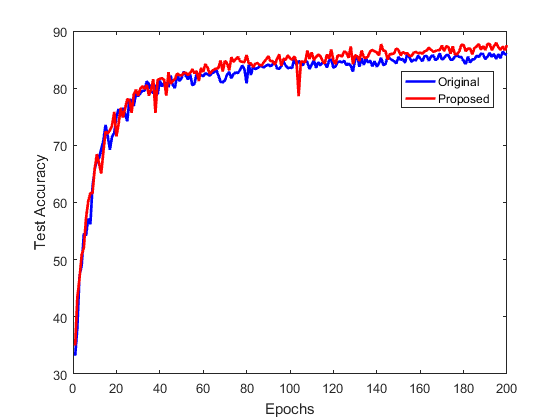}}
	\subfloat[\scriptsize EfficientNetB0-Test]{\includegraphics[width = \dimexpr\columnwidth/2\relax]{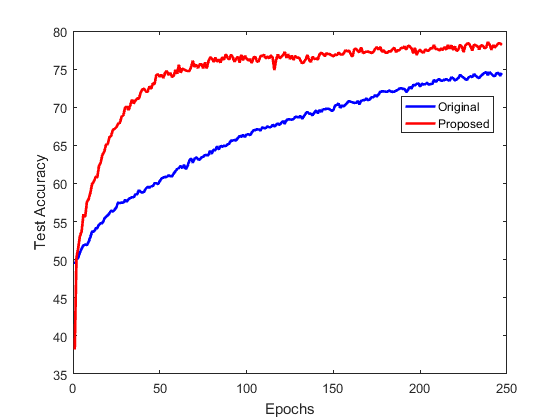}}
	\caption{Reduced-CIFAR10 -The classification accuracies are obtained by applying our proposed regularization method on four different networks while the training size of dataset is randomly reduced to 20k. Rows show training/testing results obtained from each network.}
	\label{f3}
\end{figure*}

\begin{figure*}
	
	\subfloat[\scriptsize DensNet-121-Train]{\includegraphics[width = \dimexpr\columnwidth/2\relax]{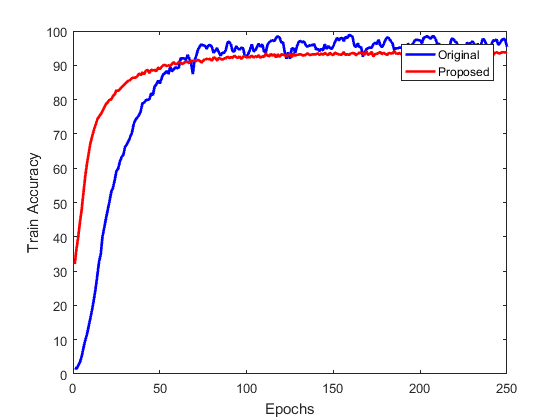}} 
	\subfloat[\scriptsize VGG13-Train]{\includegraphics[width =  \dimexpr\columnwidth/2\relax]{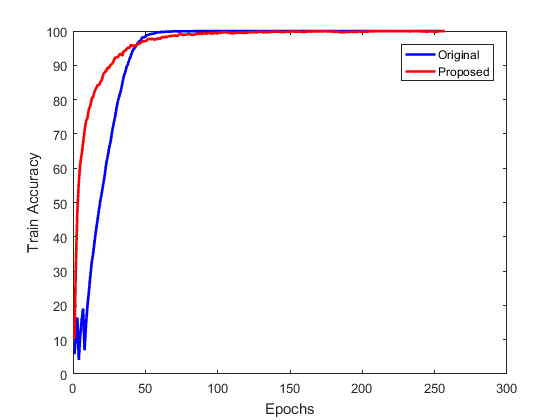}}
	\subfloat[\scriptsize ResNet152-Train]{\includegraphics[width = \dimexpr\columnwidth/2\relax]{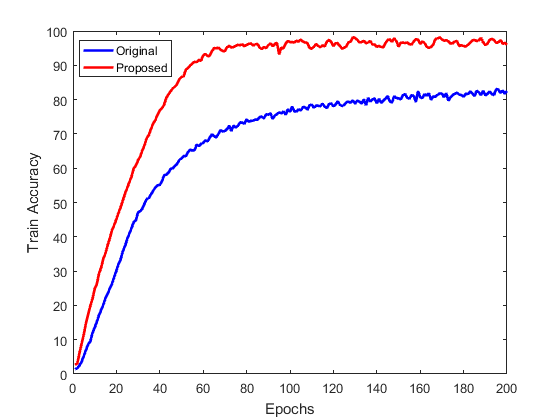}}
	\subfloat[\scriptsize EfficientNetB0-Train]{\includegraphics[width =  \dimexpr\columnwidth/2\relax]{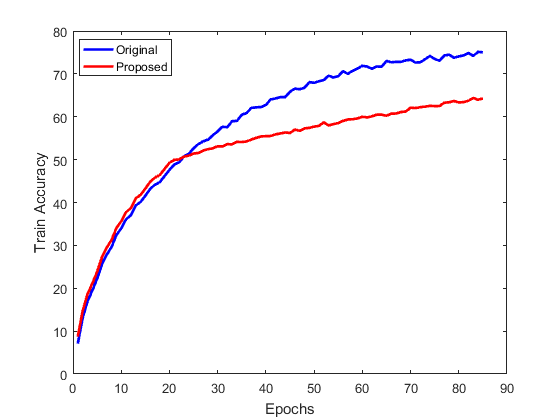}}\\
	
	\subfloat[\scriptsize DensNet-121-Test]{\includegraphics[width =  \dimexpr\columnwidth/2\relax]{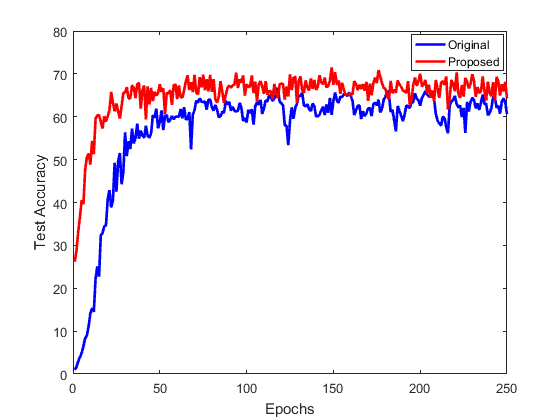}}
	\subfloat[\scriptsize VGG13-Test]{\includegraphics[width =  \dimexpr\columnwidth/2\relax]{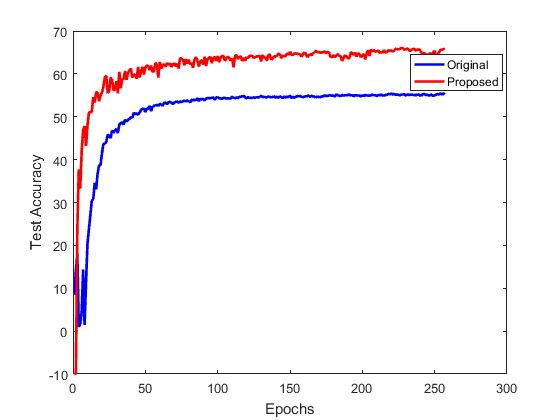}}
	\subfloat[\scriptsize ResNet152-Test]{\includegraphics[width =  \dimexpr\columnwidth/2\relax]{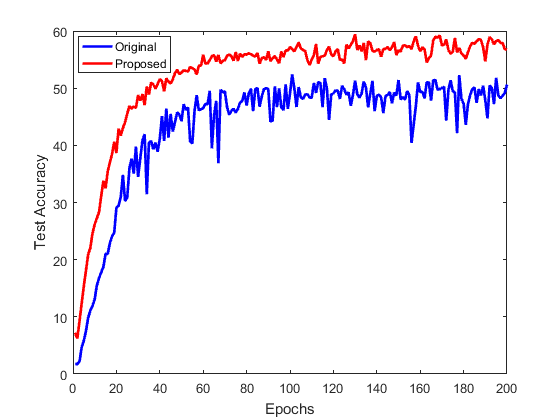}}
	\subfloat[\scriptsize EfficientNetB0-Test]{\includegraphics[width = \dimexpr\columnwidth/2\relax]{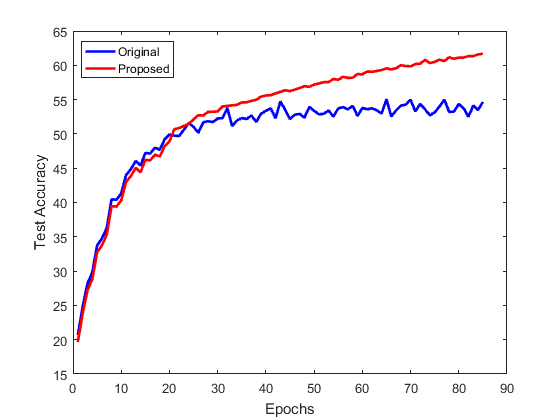}}
	\caption{Reduced-CIFAR100- The classification accuracies are obtained by applying our proposed regularization method on four different networks while the training size of dataset is randomly reduced to 20k. Rows show training/testing results obtained from each network.}
	\label{f4}
\end{figure*}

This subsection reports the results of our proposed regularization function on CIFAR-10 and CIFAR-100 and compares them to the original case of each network, i.e., the case of not using the proposed regularization. CIFAR-10 consists of $60k$ $32\times32$ color-images divided into 10 classes. Moreover, the standard training and testing sizes are $50k$ and 10k respectively, where the size of each training class is $5k$. CIFAR-100 is the same as CIFAR-10, except that it has $100$ classes, where each class has $500$ training images. Tables \ref{t1} and \ref{t2} report the classification accuracies of each model in each dataset, where the latter uses a subset of the original dataset as training data. Additionally, Figures \ref{f1} to \ref{f4} show the diagrams of train/test accuracies in accordance of Tables \ref{t1} and \ref{t2}, depicting the learning behaviors of the proposed method, and its ability to escape from local optima, e.g., sub-figures 2.(a,d). As the figures illustrate, in most cases the diagrams of proposed train accuracies are lower than those of original methods, certifying the less sensitivity of the proposed regularization to overfitting. Finally, training/testing diagrams of accuracies in Figures \ref{f1} and \ref{f2}. As the results show, applying the proposed method results in a significant improvement for each network. More importantly, Table \ref{t2} demonstrates the overfitting robustness of the proposed regularization in case of small-sample-sized problems, i.e., there are small drops in accuracies. \\
\begin{table}
	\begin{center}
		\caption{The comparison of test results on CIFAR-10 and CIFAR-100. The improvement ratio shows the amount of improvement achieved by applying the proposed regularization method. Batch-size is set to 128.}
		\label{t1}
		\resizebox{\columnwidth}{!}{%
			\begin{tabular}{lcccccc}
				\toprule
				& \multicolumn{3}{c}{ \textbf{CIFAR10}} & \multicolumn{3}{c}{ \textbf{CIFAR100}} \\\cmidrule(lr){2-4} \cmidrule(lr){5-7}
				& Original&Proposed&\shortstack{Improvement} & Original&Proposed&\shortstack{Improvement}\\ \midrule
				\textbf{Network}\\
				\;DensNet-121 &93.25 &95.63 & +\%2.55 &69.4  & 73.28 & +\%5.59 \\ 
				\;VGG-13      &92.26 & 93.66 & +\% 1.52 &67.25 & 71.3 & +\%6.02 \\ 
				\;ResNet-152  &92.71 & 95.0 & +\% 2.47 &75.67 & 77.33 & +\%2.2 \\ 
				\;EfficientNetB0&89.11 & 91.77 & +\% 2.99 &78.61 & 80.05 & +\%1.83 \\ 
				\bottomrule
			\end{tabular}
		}
	\end{center}
	
\end{table}

%%%%%%%%%%%%%%%%%%%%%%%%%%%%%%%%%%%%%%%%%%%%%%%%5

%%%%%%%%%%%%%%%%%%%%%%%%%%%%%%%%%%%%%%%%%%%%%%%%%5

\begin{table}
	
	\begin{center}
		\caption{The comparison of test results on a randomly reduced sets (20k) of CIFAR-10 and CIFAR-100. The improvement ratio shows the amount of improvement achieved by applying the proposed regularization method. Batch-size is set to 128.}
		\label{t2}
		\resizebox{\columnwidth}{!}{%
			\begin{tabular}{lcccccc}
				\toprule
				& \multicolumn{3}{c}{ \textbf{CIFAR10}} & \multicolumn{3}{c}{ \textbf{CIFAR100}} \\\cmidrule(lr){2-4} \cmidrule(lr){5-7}
				& Original&Proposed&\shortstack{Improvement} & Original&Proposed&\shortstack{Improvement}\\ \midrule
				\textbf{Network}\\
				\;DensNet-121 &89.73 & 91.95 &+\%2.47 &68.21  & 72.56 &+\%6.37 \\ 
				\;VGG-13      &88.61 & 90.62 &+\%1.70 &58.89 & 70.36 &+\%19.47 \\ 
				\;ResNet-152  &87.02 & 89.31 &+\%2.63 &62.33 & 75.42 &+\%21.0 \\ 
				\;EfficientNetB0&83.16 & 86.17 &+\%7.39 &64.3 &76.01 &+\%18.21 \\ 
				\bottomrule
			\end{tabular}
		}
	\end{center}
\end{table}
%%%%%%%%%%%%%%%%%%%%%%%%%%%%%%%%%%%%%%%%%%%%%%%%%5
%%%%%%%%%%%%%%%%%%%%%%%%%%%%%%%%%%%%%%%%%%%%%%%%%5

\subsection{ImageNet}
This experiment investigates the impact of our proposed regularization term on a very large-scale set of images. ImageNet is a large image-classification dataset with more than $1$ million annotated images divided into $1000$ classes. Table \ref{t3} reports the classification results on ImageNet and a reduced training set of ImageNet by randomly selecting 200 images from each category, therefore 200k in total. As the results verify, applying the proposed regularization method could significantly increase the performance of each method. More importantly, the improvement ratios on the reduced version of ImageNet are significantly higher than those of the full dataset, supporting the idea that the proposed method could be helpful in case of small-sample-size problems by reducing the chance of overfitting. Detailed specifications and preprocessing steps of each method are available in \textcolor{blue} {https://paperswithcode.com/sota/image-classification-on-imagenet}.

\begin{table}
	
	\begin{center}
		\caption{The comparison of top-1 test results on ImageNet and a randomly reduced set of ImageNet (200k). The improvement ratio shows the amount of improvement achieved by applying the proposed regularization method.}
		\label{t3}
		\resizebox{\columnwidth}{!}{%
			\begin{tabular}{lcccccc}
				\toprule
				& \multicolumn{3}{c}{ \textbf{ImageNet}} & \multicolumn{3}{c}{ \textbf{Reduced ImageNet}} \\\cmidrule(lr){2-4} \cmidrule(lr){5-7}
				& Original&Proposed&\shortstack{Improvement \\ Ratio} & Original&Proposed&\shortstack{Improvement \\ Ratio}\\ \midrule
				\textbf{Network}\\
				\;DensNet-121 &74.98 & 77.2 &+\%2.96  &62.9  &70.5  &+\% 12.08 \\ 
				\;VGG-13      &74.1 & 76.33 &+\%3.0   &61.25 &69.05 &+\% 12.73 \\ 
				\;ResNet-152  &78.57 & 79.8 &+\%1.56  &69.7  &75.4  &+\% 8.17 \\  
				\;EfficientNetB0&76.3 &78.15 &+\%2.42 &65.4  &69.1  &+\% 5.65 \\ 
				\bottomrule
			\end{tabular}
		}
	\end{center}
\end{table}

To have a deeper investigation on the effectiveness of the proposed method, we use class investigation maps (CAM), which is introduced in \cite{R18}, to depict class activations of each architecture on several samples of ImageNet\textquoteright s test set. To do that, we randomly select three classes of ImageNet depicted in the first row of Figure 6. Then, we calculate their CAMs using the original DensNet-121 and the one equipped with the proposed regularization (the second row shows the obtained CAMs), where both networks are trained on ImageNet. Finally and to have a better view, the third row combines the first two rows as one. From the results, it is evident that the proposed regularization forces DensNet-121 to learn discriminative features from the object of interest in each class, while the original DensNet-121 tends to memorize most areas of images. In the case of Fireweed, for instance, the equipped version of DensNet-121 with our proposed regularization uses a few number of petals to make decision, while the the original DensNet almost uses all areas of image in its decision, which eventually leads to a lower generalization.

\begin{figure*}
	\label{f7}
	\subfloat[Wading bird]{\includegraphics[width = \dimexpr\columnwidth/3*2\relax]{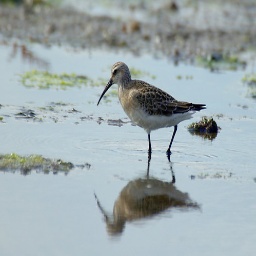}}
	\subfloat[Fireweed]{\includegraphics[width =  \dimexpr\columnwidth/3*2\relax]{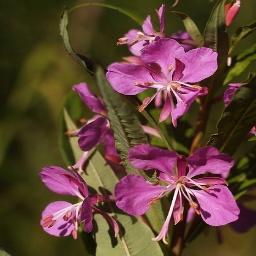}}
	\subfloat[Pouch]{\includegraphics[width = \dimexpr\columnwidth/3*2\relax]{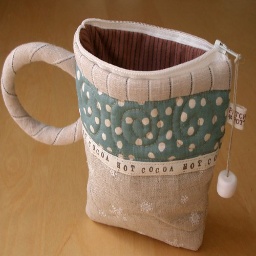}}
	
	\subfloat[proposed]{\includegraphics[width =  \dimexpr\columnwidth/3\relax]{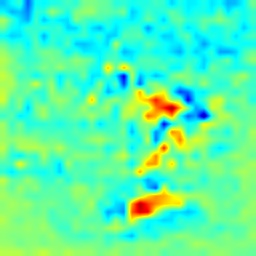}}
	\subfloat[original]{\includegraphics[width =  \dimexpr\columnwidth/3\relax]{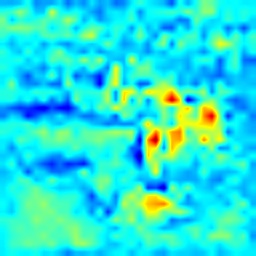}}
	\subfloat[proposed]{\includegraphics[width =  \dimexpr\columnwidth/3\relax]{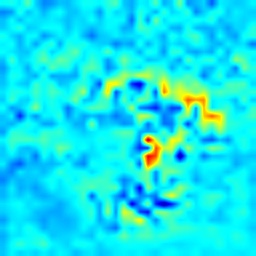}}
	\subfloat[original]{\includegraphics[width =  \dimexpr\columnwidth/3\relax]{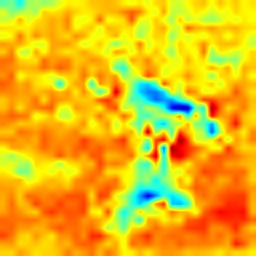}}
	\subfloat[proposed]{\includegraphics[width =  \dimexpr\columnwidth/3\relax]{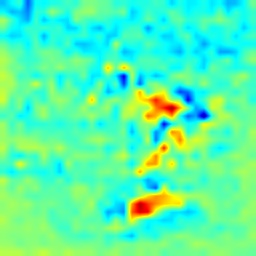}}
	\subfloat[original]{\includegraphics[width =  \dimexpr\columnwidth/3\relax]{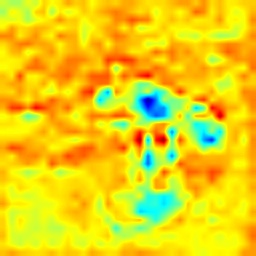}}
	
	\subfloat[proposed]{\includegraphics[width =  \dimexpr\columnwidth/3\relax]{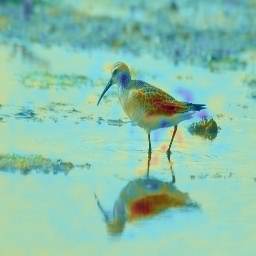}}
	\subfloat[original]{\includegraphics[width =  \dimexpr\columnwidth/3\relax]{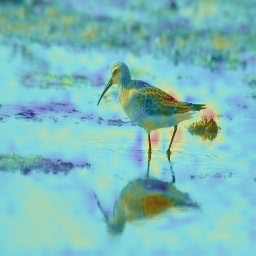}}
	\subfloat[proposed]{\includegraphics[width =  \dimexpr\columnwidth/3\relax]{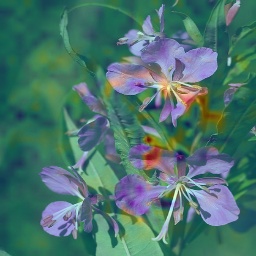}}
	\subfloat[original]{\includegraphics[width =  \dimexpr\columnwidth/3\relax]{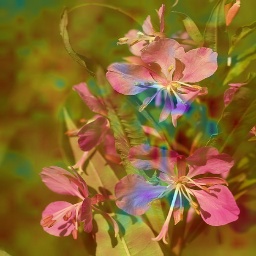}}
	\subfloat[proposed]{\includegraphics[width =  \dimexpr\columnwidth/3\relax]{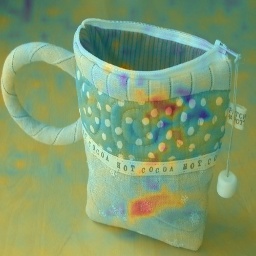}}
	\subfloat[original]{\includegraphics[width =  \dimexpr\columnwidth/3\relax]{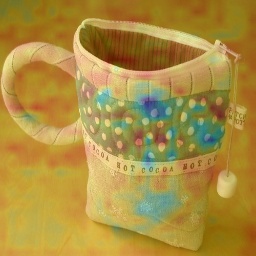}}
	
	\caption{The results of Class Activation Mapping (CAM) of DensNet-121 on several images taken from ImageNet. The first row shows original images, and the second row depicts the obtained CAMs with and without the proposed regularization. The third row combines the previous rows to have a better view of CAMs. As the results show, the CAMs of the proposed regularization highlight less but more discriminative areas of the image, e.g., the wings of the bird, which is very discriminative, are highlighted even on the water, while the CAMs of the original network tend to highlight more but less discriminative areas of the image, e.g, look at sub-figures (g,i).}
	
\end{figure*}

\subsection{Comparison with $\ell_2-norm$ regularizer}
\label{comp}
\begin{figure*}
	
	\subfloat[\centering \scriptsize Train-Acc with Dropout - Epochs 1-200]{\includegraphics[width = \dimexpr\columnwidth/2\relax]{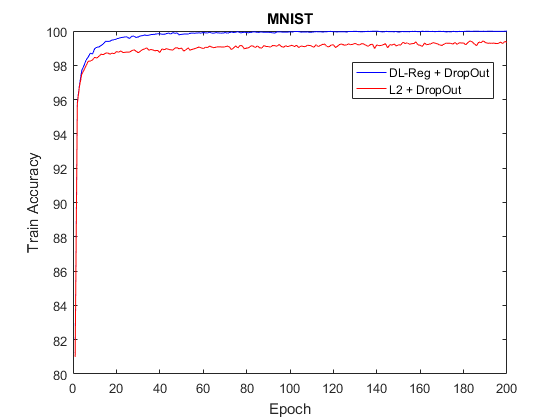}}
	\subfloat[\centering \scriptsize Train-Acc with Dropout - Epochs 800-1200]{\includegraphics[width = \dimexpr\columnwidth/2\relax]{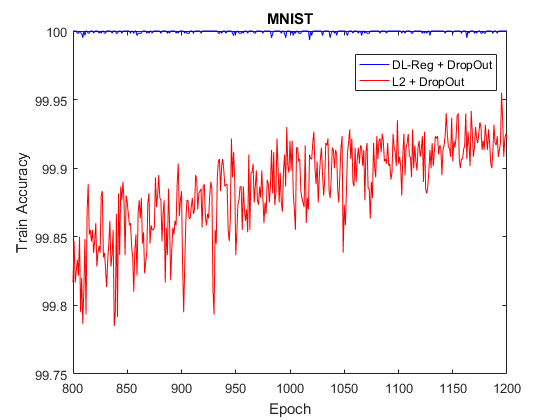}}
	\subfloat[\centering \scriptsize Train-Acc without Dropout - Epochs 1-200]{\includegraphics[width =  \dimexpr\columnwidth/2\relax]{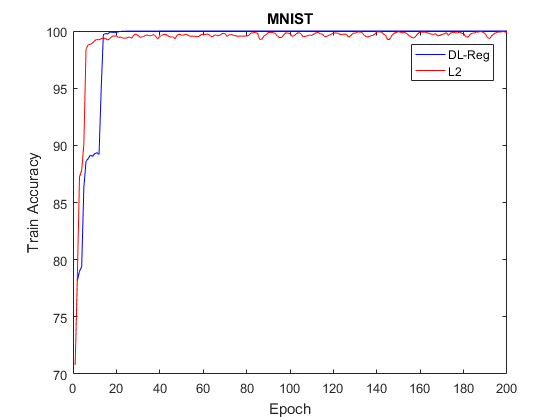}}
	\subfloat[\centering \scriptsize Train-Acc without Dropout - Epochs 800-1200]{\includegraphics[width =  \dimexpr\columnwidth/2\relax]{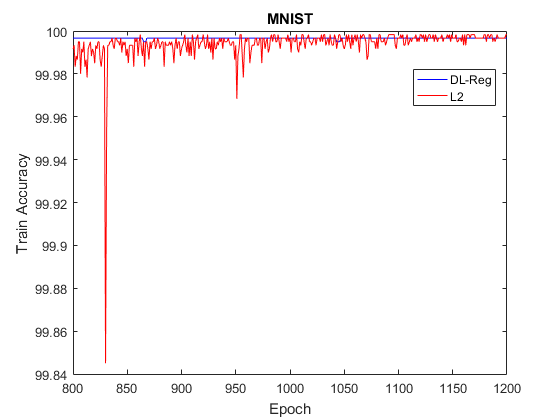}}

	%----------------------------------------------------------------------------------------------
	\subfloat[\centering \scriptsize Test-Acc with Dropout - Epochs 1-200]{\includegraphics[width = \dimexpr\columnwidth/2\relax]{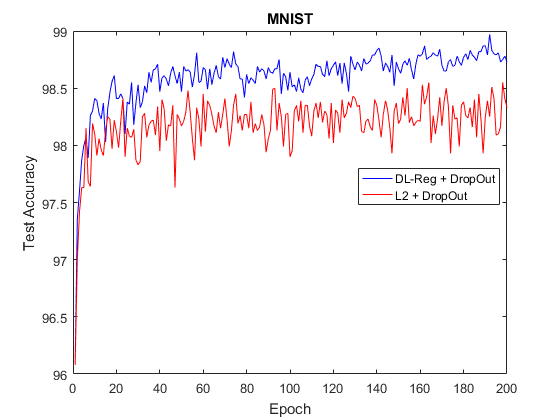}}
	\subfloat[\centering \scriptsize Test-Acc with Dropout - Epochs 800-1200]{\includegraphics[width = \dimexpr\columnwidth/2\relax]{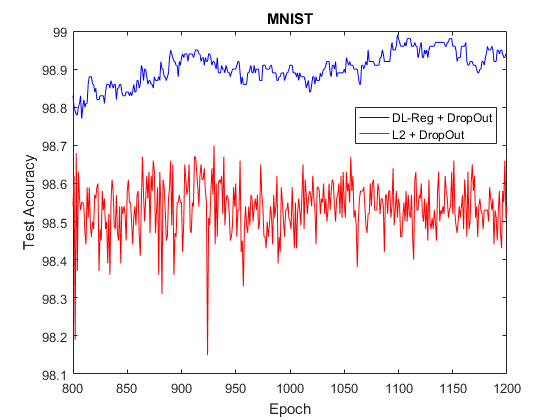}}
	\subfloat[\centering \scriptsize Test-Acc without Dropout - Epochs 1-200]{\includegraphics[width =  \dimexpr\columnwidth/2\relax]{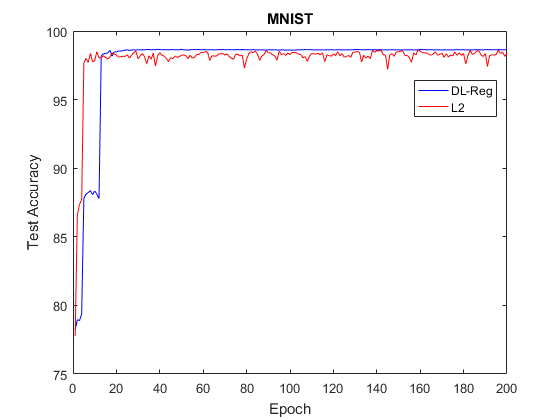}}
	\subfloat[\centering \scriptsize Test-Acc without Dropout - Epochs 800-1200]{\includegraphics[width =  \dimexpr\columnwidth/2\relax]{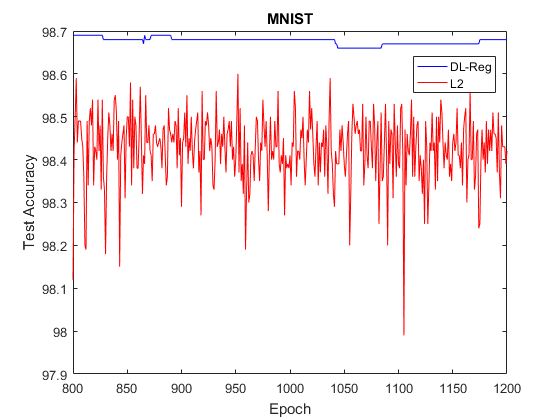}}
	
	%-----------------------------------------------------------------------------------------------
	\subfloat[\centering \scriptsize Train-loss with Dropout - Epochs 1-200]{\includegraphics[width = \dimexpr\columnwidth/2\relax]{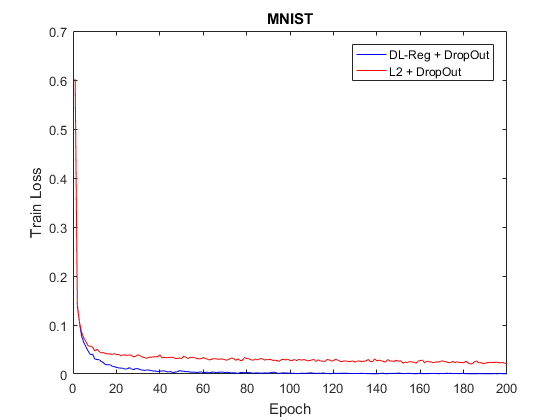}}
	\subfloat[\centering \scriptsize Train-loss with Dropout - Epochs 800-1200]{\includegraphics[width = \dimexpr\columnwidth/2\relax]{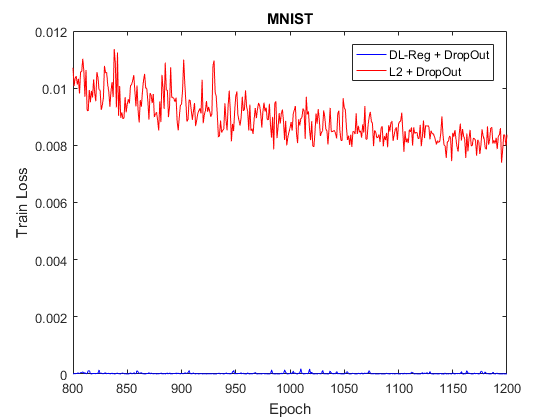}}
	\subfloat[\centering \scriptsize Train-loss without Dropout - Epochs 1-200]{\includegraphics[width =  \dimexpr\columnwidth/2\relax]{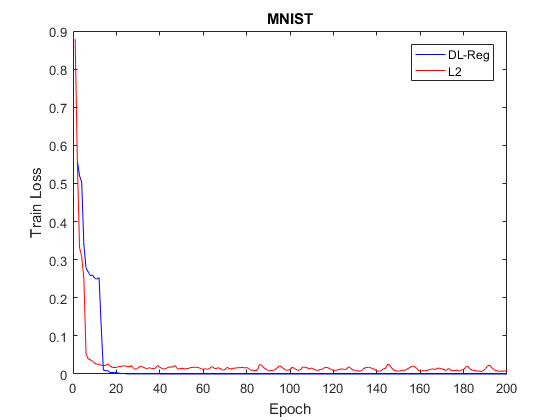}}
	\subfloat[\centering \scriptsize Train-loss without Dropout - Epochs 800-1200]{\includegraphics[width =  \dimexpr\columnwidth/2\relax]{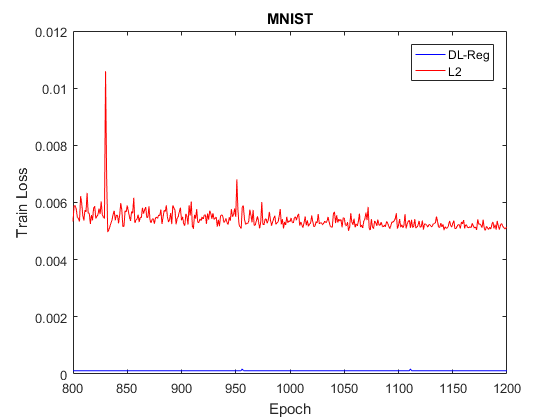}}

	\caption{A comparison between the traditional $\ell_2$ regularizer and the proposed DL-Reg in two scenarios including with (first two columns) and without (last two columns) Dropout layers on the MNIST dataset. The network structure and all other common settings are the same for both methods. To better visualize the training behaviour of models, only the results of the first 200 epochs and the last 400 epochs are shown. In both scenarios, DL-Reg outperforms $\ell_2$ regularizer. Moreover, DL-Reg shows faster convergence and more stable behaviours. }
	\label{f8}
\end{figure*}

This subsection conducts several experiments on the MNIST dataset to compare the performance of the proposed regularization technique (DL-Reg) and the well-known $\ell_2-norm$. The reason for selecting $\ell_2$ regularizer is that it belongs to the same category (see Section \ref{intro}) as DL-Reg; hence, the comparison is fair. In all experiments, we use the same parameter settings including randomness, train/test size, batch-size, learning rate, max-epoch, and every other setting. Table \ref{t5} describes a list of such parameters along with their assigned values. Moreover, we use the same network structure, consisting of three sequential hidden layers (1024, 1024, 2048) with ReLUs and with/without Dropout rates of $0.2$ for the input-layer and $0.5$ for the other layers. Figure \ref{f8} depicts the obtained per-epoch results in terms of train and test accuracies as well as train losses. To have a better view over the results, we only depict the results of the first 200 epochs and the last 400 epochs of the learning procedure. Therefore, it becomes easier to compare the learning behaviours of models at the beginning and the end of training. Additionally, the final test accuracy of each strategy is reported in Table \ref{t4}. 

As it is depicted in Figure \ref{f8}, in all cases the proposed regularization method achieves higher accuracy in both test and train phases, and a lower value of training loss. More precisely, the proposed DL-Reg shows a superior behaviour to $\ell_2$ regularizer in both cases of with and without Dropout layers. The convergence speed is another significant implication of the proposed method. We can observe that DL-Reg shows even a faster rate of convergence and a more stable behavior compared to $\ell_2$ regularizer. That is to say that DL-Reg can successfully reduce the nonlinearity of deep networks by implicitly forcing the neurons of the networks to behave as linear as necessary. 

Another interesting observation by investigating Figure \ref{f8}(c) is the fact that $\ell_2$ regularizer performs better at the first epochs of the training; however, after a certain number of epochs, DL-Reg reveals its generalization power and performs superior to $\ell_2$ regularizer. %This observation could be justified by considering the fact that the proposed regularizer is based on a linear technique, and linear techniques are generally simpler, i.e., lower learning ability and a higher rate of generalization compared to the nonlinear ones; hence, at the beginning steps of learning, $L_2$ benefits from its nonlinear nature and learns faster than the proposed regularizer. On the other hand, the linear methods benefit from a higher rate of generalization, and, therefore, they are less prone to be trapped into a local optimum compared to the case of nonlinear methods. Hence, after a while, we can see that DL-Reg prevents the network from converging into a local optimum and performs better, while $L_2$ tends to converge to the nearest locale optima. This observation, however, does not happen in sub-figure (a) because this scenario uses DropOuts, and DropOuts generally lowerizes the nonlinearity of models to a considerable extent. Therefor, DL-Reg performs well even at the very beginning epochs.

\begin{table}[ht]
    \caption{Comparison of $\ell_2$ and the proposed DL-Reg regularization methods on MNIST dataset.}
    
    %\resizebox{0.7\columnwidth}{!}{%
        %\tiny
        \centering
        \begin{tabular}{l c}
        \hline\hline 
         Method &  Test Classification Accuracy \\ 
        \hline
        $\ell_2$ regularizer & 98.38  \\
        Dropout$+\ell_2$ regularizer & 98.47  \\
        DL-Reg (proposed) & 98.69\\
        Dropout$+$DL-Reg (proposed) & \textbf{98.94}  \\
        \hline
        \end{tabular}
   % }
    \label{t4}
\end{table}

\begin{table}[ht]
    \caption{The hyper-parameters of the fully-connected networks ($784 \rightarrow 1024 \rightarrow 1024 \rightarrow 2048 \rightarrow s10$) trained on MNIST dataset, (Subsection \ref{comp}).}
    
    %\resizebox{0.7\columnwidth}{!}{%
        %\tiny
        \centering
        \begin{tabular}{l c c}
        \hline\hline 
         Parameter &  Value \\ 
        \hline
        Learning rate (lr) & $0.1$  \\
        Decay rate for lr & 0.96  \\
        lr-scheduler & Exponential \\
        Frequency of lr-scheduler & every 30 epochs\\
        Optimizer & SGD \\
        Momentum & $0.9$ \\
        Loss function & Cross-Entropy\\
        Batch size & $256$ \\
        Data pre-processing & None \\
        Regularization factor for DL-Reg & $1e-12$\\
        Regularization factor for $\ell_2$ regularizer & $5e-4$\\
        \hline
        \end{tabular}
   % }
    \label{t5}
\end{table}

\section{Discussion and Future Work}
\label{sec5}
There is always a trade-off between the amount of data used for training and the depth of the model on one side, and the model's complexity in terms of memory and time on another side. The proposed regularization technique forces the network to behave as linear as possible. That is, it limits the network to learn a highly nonlinear function while preserving its prediction's ability. This limitation enables the network to learn discriminative features. This ability is clearly visible in the obtained results depicted in Figure 6. In the case of Wading bird, for instance, the proposed method uses the wings' pattern of the bird for detecting this object, which obviously provides enough discrimination. It is worthy to note that the proposed method even detects the wings' reflection on the water, which is incredible.

The parameter of regularization factor, i.e., $\gamma$, plays an essential role in the performance of DL-Reg. If $\gamma$ increases, then the learning ability of the network reduces, causing the network to entirely behave like a linear regression. In contrast, if $\gamma$ approaches zero, then there would be no more regularization/generalization impact in the learning procedure. Therefore, the parameter$\gamma$ should be chosen carefully in every learning problem.

Finally, the main implications of the proposed regularization method are summarized as follows:
\begin{enumerate}
	\item DL-Reg provides a better generalization in practice and learns discriminative features
	\item The convergence speed of the proposed DL-Reg is fast; however, it depends to the value of regularization factor
	\item The computational cost of DL-Reg is negligible
	\item the proposed regularization method is easy to implement and can be added to any network. In other words, it is independent of the choice of loss-function.
\end{enumerate}

One of the promising areas of future work could be investigating the effects of adding the linearity restriction to each layer of the network, jointly or separately. %Moreover, the combination of the proposed method and other regularizers, e.g., Dropout,  could be also considered as a promising future work.

\section{Conclusion}
\label{sec6}
This paper proposes a linear technique named as DL-Reg for regularizing the family of deep neural networks. As such deep networks tend to learn highly nonlinear functions, DL-Reg forces the final network to behave as linear as possible and, at the same time, as nonlinear as necessary. A series of various experiments along with a comparison with the traditional $\ell_2$ regularizer is conducted, and the obtained results show great improvements in classification performances of several state-of-the-art methods. We have also shown that DL-Reg is able to extract discriminative features while avoiding unnecessary and less discriminative ones. This behavior enables the final network to avoid overfitting. Moreover, the proposed method is easy to implement and increases the learning/convergence speed.

\ifCLASSOPTIONcaptionsoff
  \newpage
\fi

\bibliographystyle{ieee_fullname}
\bibliography{egbib}

\begin{IEEEbiography}[{\includegraphics[width=1in,height=1.25in,clip,keepaspectratio]{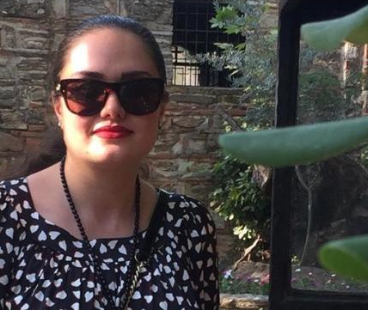}}]{Maryam Dialameh}
received her A.S. degree (2009) in computer engineering from University of Applied Science and Technology of Shiraz in Iran. She then received her B.S. degree (2012) in computer engineering and her M.S. degree (2017) in artificial intelligence from Azad University and Shiraz University in Iran respectively. Her research interests include Machine Learning, Deep Learning, and Computer Vision. She has also published several national and international papers in these areas and beyond.
\end{IEEEbiography}
\begin{IEEEbiography}[{\includegraphics[width=1in,height=1.25in,clip,keepaspectratio]{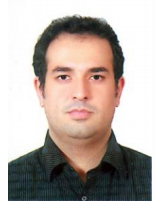}}]{Ali Hamzeh}

received his B.S. and M.Sc. degrees in computer engineering and artificial
intelligence from Shiraz University in Iran. Then he received his Ph.D. degrees in artificial
intelligence from Iran University of Science and Technology, 2007. He has been an assistant
professor of artificial intelligence at Shiraz University from 2007. His research interests include
evolutionary computation, machine learning and social networks. Also, he has published over
150 peer-refereed articles in these areas.
\end{IEEEbiography}

\begin{IEEEbiography}[{\includegraphics[width=1in,height=1.25in,clip,keepaspectratio]{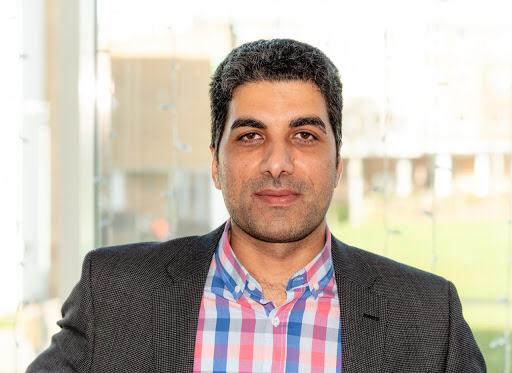}}]{Hossein Rahmani}
received the B.Sc. degree in computer software engineering from the Isfahan University of Technology, Isfahan, Iran, in 2004, the M.Sc degree in software engineering from Shahid Beheshti University , Tehran, Iran, in 2010, and the Ph.D. degree from The University of Western Australia, in 2016. He has published several papers in top conferences and journals such as CVPR, ICCV, ECCV, IJCV, IEEE TIP, IEEE TPAMI. He has received several research grants from EPSRC, Innovate UK and EU with combined funding of over \$2.5 million. He is currently an Associate Professor (Lecturer) with the School of Computing and Communications, Lancaster University. Before that, he was a Research Fellow at the School of Computer Science and Software Engineering, The University of Western Australia. His research interests include computer vision and machine learning.
\end{IEEEbiography}

%\newpage



%\vfill



\end{document}